\documentclass[letterpaper, 10 pt, conference]{ieeeconf}  

\IEEEoverridecommandlockouts                              

\usepackage{graphics} 
\usepackage{epsfig} 
\usepackage{mathptmx} 
\usepackage{times} 
\usepackage{amsmath} 
\usepackage{amssymb}  
\usepackage{float}
\usepackage{booktabs}
\usepackage{graphicx}
\usepackage[hyphens]{url}

\title{\LARGE \bf
CamVox: A Low-cost and Accurate Lidar-assisted Visual SLAM System
}

\author{Yuewen Zhu, Chunran Zheng, Chongjian Yuan, Xu Huang and Xiaoping Hong
\thanks{*Research is supported by SUSTech start-up fund and DJI-SUSTech joint lab.}
\thanks{All authors are with the School of System Design and Intelligent Manufacturing, Southern University of Science and Technology, China 
{\tt\small \{zhuyw2019,zhengcr,ycj2020,huangx2020\}@mail.
sustech.edu.cn, hongxp@sustech.edu.cn}.
        }
}

\begin{document}

\maketitle
\thispagestyle{empty}
\pagestyle{empty}

\begin{abstract}

Combining lidar in camera-based simultaneous localization and mapping (SLAM) is an effective method in improving overall accuracy, especially 
at a large scale outdoor scenario. Recent development of low-cost lidars (e.g. Livox lidar) enable us to explore such SLAM systems with lower 
budget and higher performance. In this paper we propose CamVox by adapting Livox lidars into visual SLAM (ORB-SLAM2) by exploring the lidars’ unique features. Based on the non-repeating nature of Livox lidars, we propose an automatic lidar-camera calibration 
method that will work in uncontrolled scenes. The long depth detection range also benefit a more efficient mapping. Comparison of CamVox with 
visual SLAM (VINS-mono) and lidar SLAM (LOAM) are evaluated on the same dataset to demonstrate the performance. We open sourced our hardware, code and 
dataset on GitHub\footnote{\url{https://github.com/ISEE-Technology/CamVox}}.

\end{abstract}

\section{INTRODUCTION}

Simultaneous localization and mapping (SLAM) is a key technique in autonomous robots with growing attention from academia and industry. The advent of low-cost sensors accelerated significantly the SLAM development. For example, cameras provided rich angular and color information and enabled the development of visual SLAM such as ORB-SLAM \cite{mur2015orb}. Following that the inertial measurement units (IMU) start to drop price thanks to their massive adoption in smart phone industry, and the utilization in SLAM becomes straightforward to gain additional modality and performance such as in VINS-mono \cite{qin2018vins} and ORB-SLAM3 \cite{campos2020orb}. Among the additional sensors, depth sensors (stereo camera, RGB-D camera) provide direct depth measurement and enables accurate performance in SLAM applications such as ORB-SLAM2 \cite{mur2017orb}. lidar, as the high-end depth sensor, provides long range outdoor capability, accurate measurements and system robustness, and has been widely adopted in more demanding applications such as autonomous driving \cite{urmson2008autonomous}, but also typically come with a hefty price tag. As the autonomous industry progresses, recently many new technology developments have enabled commercialization of low-cost lidars, e.g. Ouster and Livox lidars. Featuring a non-repeating scanning pattern, Livox lidars pose unique advantageous in low-cost lidar-assisted SLAM system. We in this paper present the first Livox lidar assisted visual SLAM system (CamVox) with superior and real-time performance. Our CamVox SLAM built upon the state-of-the-art ORB-SLAM2 by using Livox lidar as the depth sensor, with the following new features:

\begin{enumerate}

\item A preprocessing step fusing lidar and camera input. Careful time synchronization is performed that the distortions in the non-repeating scanned lidar points are corrected by the IMU data and are transformed to camera frames. 
\item The accuracy and range of the lidar point cloud is superior compared to other depth cameras, and the proposed SLAM system can perform large-scale mapping with higher efficiency and can robustly operate in an outdoor strong-sunlight environment. 
\item We utilized the non-repeating scanning feature of Livox lidar to perform automatic calibrations between the camera and lidar at uncontrolled scenes.
\item CamVox performance is evaluated against some main stream frameworks and exhibits very accurate trajectory results. We also open-sourced the hardware, code and dataset of this work hoping to provide an out-of-the-box CamVox (Camera + Livox lidar) SLAM solution.
\end{enumerate}

\begin{figure}[h]
    \centering
    \includegraphics[width=\linewidth]{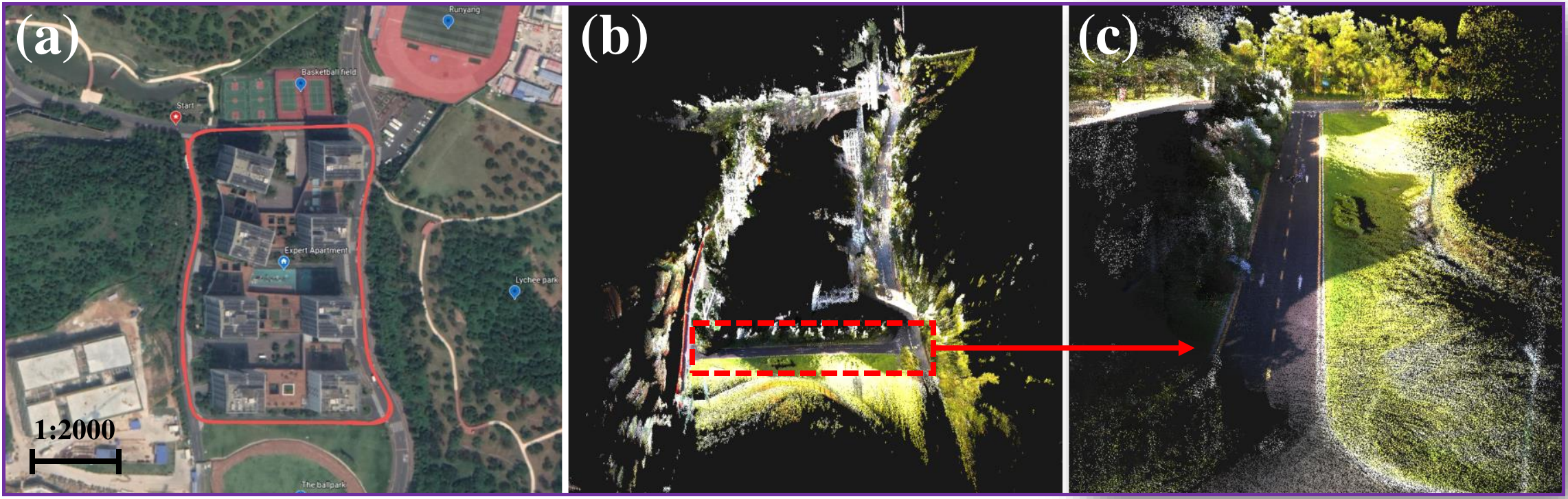}
    \caption{Example of CamVox performance. (a) the trajectory of the robot from CamVox; (b) the dense RGBD map from CamVox; (c) a magnified and rotated part of (b) demonstrating high point cloud quality.}
    \label{fig:example}
\end{figure}

Fig. \ref{fig:example} shows an example trajectory and map construction from CamVox. The scale of the map is about 200 meters on the long side. The rest of this paper is structured as the following. Related work is reviewed in Section II. The CamVox system hardware and software framework is described in detail in Section III. The results and evaluation are presented in Section IV. Finally, we conclude and remark the outlook in Section V.

\section{Related work}

Due to the rich angular resolution and informative color information, cameras could provide surprisingly good localization and mapping performance through a simple projection model and bundle adjustment. One of the most famous SLAM systems with monocular camera is ORB-SLAM \cite{mur2015orb}. ORB-SLAM tracks the object by extract the ORB features in the image and use loop-closure detection to optimize the map and pose globally, which is usually fast and stable. However, it cannot accurately recover the real scale factor since an absolute depth scale is unknown to a camera.

To fix the issue that monocular camera doesn’t recover real-world scale, Mur et al. proposed an improvement of ORB-SLAM named ORB-SLAM2 \cite{mur2017orb}, adding the support of stereo camera and RGBD camera for depth estimation. However, there are drawbacks with both of these cameras, especially in estimating the outdoor objects with long depths. The stereo camera requires a long baseline for accurate long-depth estimation, which is usually limited in real world scenarios. Additionally, the calibration between the two cameras is susceptible to mechanical changes and will adversely influence the long-depth estimation accuracy. The RGBD camera is usually susceptible to sun light with a finite range of less than 10 meters typically. 
Fusing camera and IMU is another common solution, because camera can partially correct IMU integral drift, calibrate IMU bias while IMU can overcome the scale ambiguity of monocular system. Qin et al. proposed Visual-Inertial Monocular SLAM (VINS) \cite{qin2018vins}, which is an excellent camera and IMU fusion system. Similarly, Campos et al. extended ORB-SLAM2 by fusing camera and IMU measurement and proposed ORB-SLAM3 \cite{campos2020orb}. However, consumer-grade IMU only works well in relatively low precision and suffers from bias, noise and drift while high-end IMU is prohibitively costly.

Lidar on the other hand, provides a direct spatial measurement. Lidar SLAM framework has been developed. One pioneering work is LOAM \cite{zhang2014loam}. Comparing to visual SLAM, it is more robust in a dynamic environment, due to the accurate depth estimation from lidar point cloud. However, due to the sophisticated scanning mechanism, lidar cannot provide as much angular resolution as camera does, so that it is prone to fail in environments with less prominent structures like tunnels or hallways. It also lacks loop-closure detection which make the algorithm focus only on local accuracy without global pose and map optimization. To add loop-closure detection in LOAM, Shan et al. \cite{shan2018lego} proposed an enhanced LOAM algorithm LeGO-LOAM. Comparing to LOAM, LeGO-LOAM improves feature extraction with segmentation and clustering for efficiency improvement, and adds loop-closure detection for long run drift reduction. 
Combining lidar and camera in a SLAM framework become an ideal solution. While obtaining the point cloud with accurate depth information, it could make use of the high angular and color information from camera. Zhang et al. proposed VLOAM \cite{zhang2015visual}, which fuse monocular camera and lidar loosely. Similar to LOAM, the estimation is claimed to be accurate enough and no loop-closure is needed. Shin et al. also tried to combine monocular camera and lidar together using direct method rather than feature points to estimate the pose. In addition, it tightly couples the visual data and point clouds, and output the estimated pose. Shao et al. \cite{shao2019stereo} went further fusing the stereo camera, IMU and lidar together. They demonstrated a good performance in outdoor real-scale scenes taking advantage of stereo Visual-Inertial Odometry (VIO) loop closure. VIO and lidar mapping are loosely coupled without further optimization at back-end. It is also limited by its complexity and cost.

\begin{figure}[h]
    \centering
    \includegraphics[width=\linewidth]{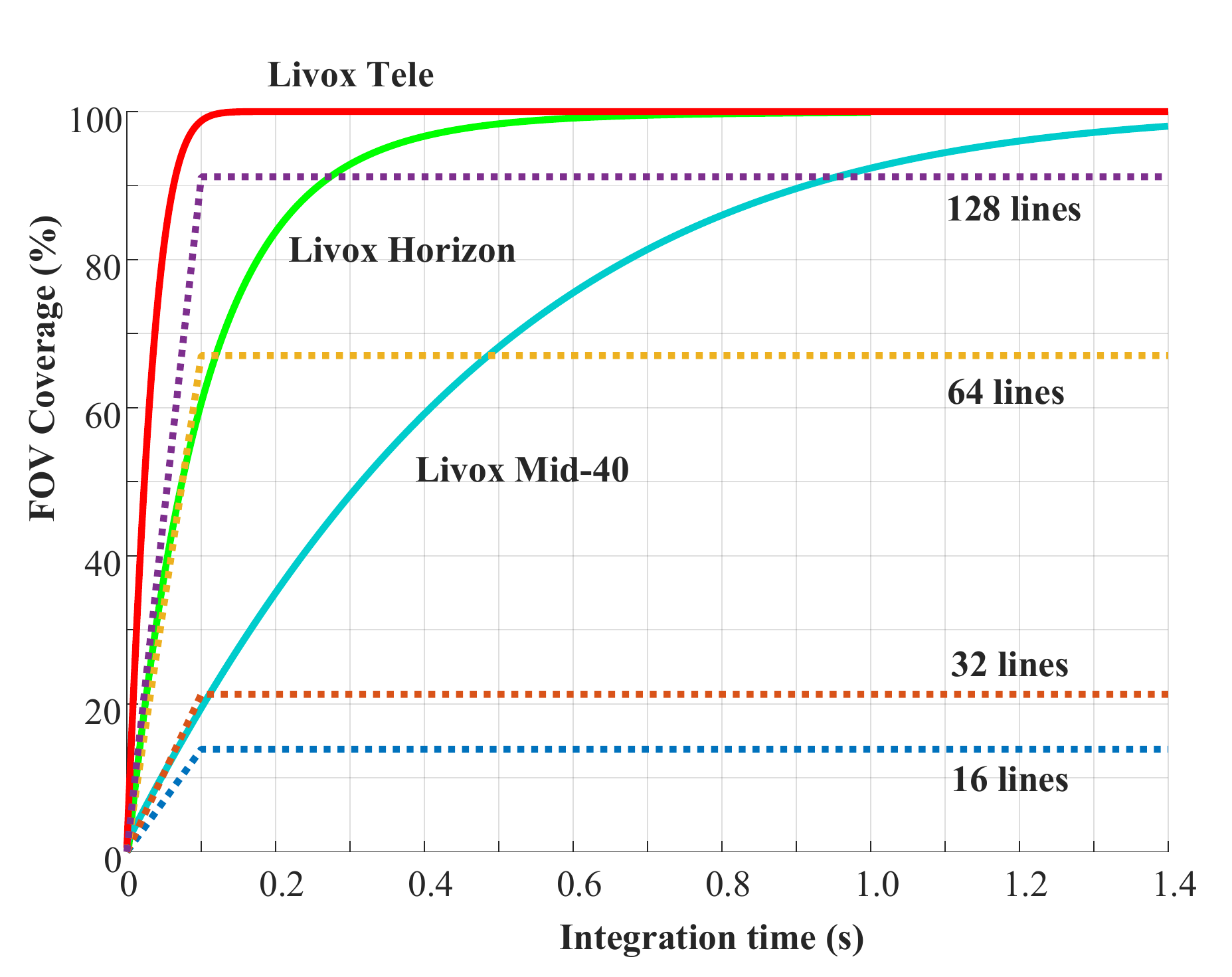}
    \caption{Point cloud density of three Livox lidar models compared to traditional lidars as a function of integration time.}
    \label{fig:fov}
\end{figure}

Lidar was typically too costly to be useful in practical applications. Fortunately, Livox unveiled a new type of lidars based on prism scanning \cite{liu2020low}. Due to the new scanning method, the cost can be significantly lowered to enable massive adoption. Furthermore, this new scanning method allows non-repeating scanning patterns\footnote{\url{https://www.livoxtech.com/3296f540ecf5458a8829e01cf429798e/assets/horizon/04.mp4}}, ideal for acquiring a relatively high-definition point cloud when accumulated (Fig. \ref{fig:fov}). Even for 100 ms accumulation, the density of Livox Horizon is already as high as 64 lines and continue to increase. This feature can be extremely beneficial in calibrating the lidar and camera, where the traditional multiline lidar lacks the precision for space between the lines. The prism design also features a maximal optical aperture for signal reception and allows long range detection. For example, the Livox Horizon could detect up to 260 m under strong sunlight\footnote{\url{https://www.livoxtech.com/horizon/specs}}. With such superior cost and performance, Lin et al. proposed Livox-LOAM \cite{lin2020loam}, which is an enhancement of LOAM adapted to Livox Mid. Based on this, Livox also released a LOAM framework for Livox Horizon, named livox\_horizon\_loam\footnote{\url{https://github.com/Livox-SDK/livox_horizon_loam}}.

Because of the long range detection and high accuracy, extrinsic parameter calibration between the camera and the lidar become a more important consideration. In \cite{levinson2013automatic}, the proposed solutions to the lidar-camera can be clustered in two ways. The first one is whether the calibration process needs a calibration target, while the second is whether the calibration can work without human intervention. During these years, many calibration techniques are based on fixed calibration target or manual effort, like \cite{unnikrishnan2005fast} and \cite{geiger2012automatic}. In \cite{pandey2012automatic}, Pandey et al. use Cramer-Rao-Lower-Bound (CRLB) to prove the existence of calibration parameters and estimate them by calculating the minimum variance unbiased (MVUB) estimator. Iyer et al. proposed a network called CalibNet, a geometrically supervised deep network to estimate the transformation between lidar and camera in \cite{iyer2018calibnet}. No specific scene is required for the above two methods. In addition, Levinson et al. proposed an online calibration method in \cite{levinson2013automatic}, in which they claimed that such a method can calibrate the lidar and camera in real time and it was suitable in any scenes. But so far, the calibration still remains as a challenge task and there is no open-source algorithm for calibrating lidar and camera in uncontrolled scene. Livox lidar’s non-repeating scanning pattern could provided a much easier solution as we will demonstrate.

\section{CamVox framework}
The proposed CamVox is based on ORB-SLAM2 (RGBD model) with separate RGBD input preprocessing and automatic calibration methods at uncontrolled scenes. The framework utilizes lidar-assisted visual keyframes to generate local mapping, and exhibits high robustness thanks to the back-end lightweight pose-graph optimization at various levels of bundle adjustment (BA) and loop closure from ORB-SLAM2. 

In the original ORB-SLAM2, keypoints are classified in two categories, close and far, where close points are those with high certainty in depth and can be used for scale, translation and rotation estimations while the far points are only used for rotation estimation and hence less informative. With the dense, long range and accurate points obtained from Livox lidars fused with camera image, many more close points could be assigned than traditional RGBD cameras (due to detection range) or stereo vision cameras (due to limited baseline). As a result, the advantages from both the camera (high angular resolution for ORB detection and tracking) and lidar (long range and accurate depth measurement) can be exploited in a tightly coupled manner.

\begin{figure}[h]
    \centering
    \includegraphics[width=\linewidth]{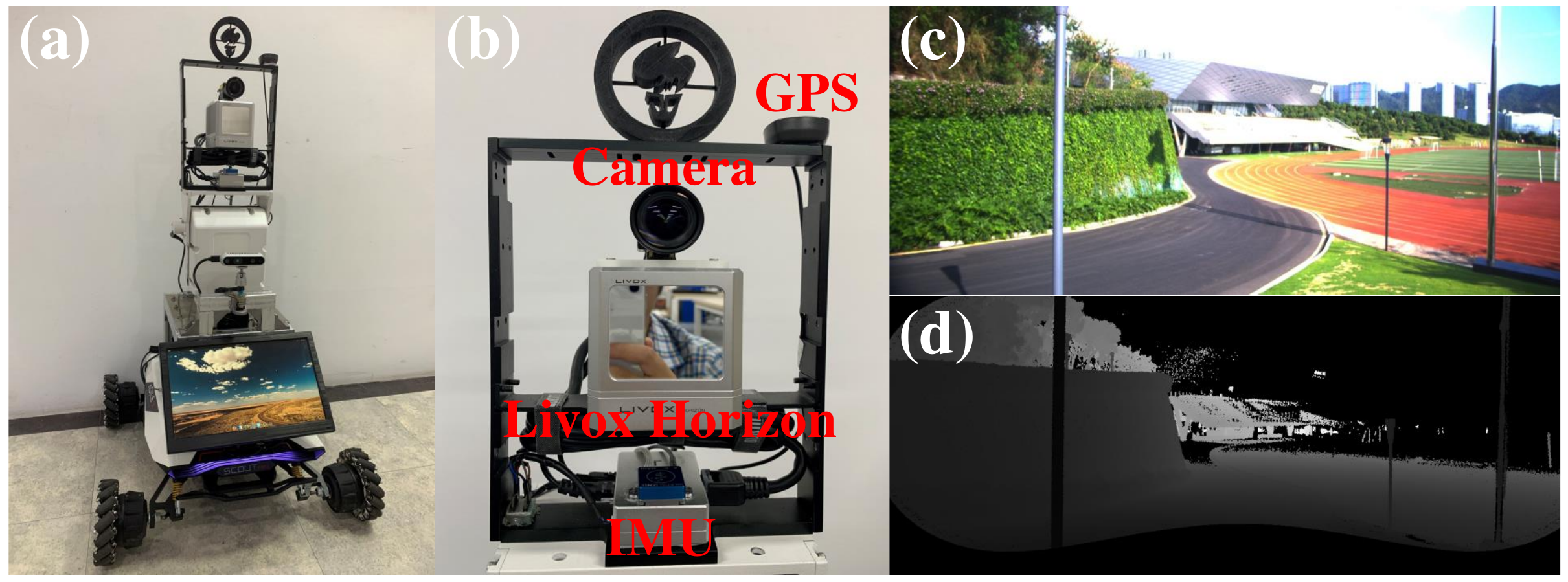}
    \caption{(a)  The complete robot platform. CamVox hardware is mounted on top of this robot. An additional RGBD camera is mounted for comparison. (b) CamVox hardware close-up including a camera, Livox Horizon, and IMU. Additional GPS/RTK is used for ground truth estimation. (c-d) an example of acquired RGB image, and depth image from lidar point cloud (colored in depth).}
    \label{fig:platform}
\end{figure}

\subsection{Hardware and software}
The CamVox hardware includes a MV-CE060-10UC rolling shutter camera, a Livox Horizon lidar and an IMU (Inertial Sense $\mu$INS). Additional GPS-RTK (Inertial Sense $\mu$INS) is used for ground truth estimation. Hard synchronization is performed with all of these sensors by a trigger signal of 10 Hz. The camera output at each trigger signal (10 Hz). The lidar keeps a clock (synced with GPS-RTK) and continuously outputs the scanned point with an accurate timestamp. In the meantime, the IMU outputs at a frequency of 200 Hz synced with the trigger. Data from the GPS-RTK is also recorded for ground truth comparison. An Intel Realsense D435 RGBD camera is mounted for comparison. The whole system mounts on a moving robot platform (Agile X Scout mini). It is noted that the sale price of the Livox Horizon lidar (800 USD) is significantly lower than other similar performance lidars (10k – 80k USD) and this allows building the complete hardware system within a reasonable budget.

\begin{figure}[h]
    \centering
    \includegraphics[width=\linewidth]{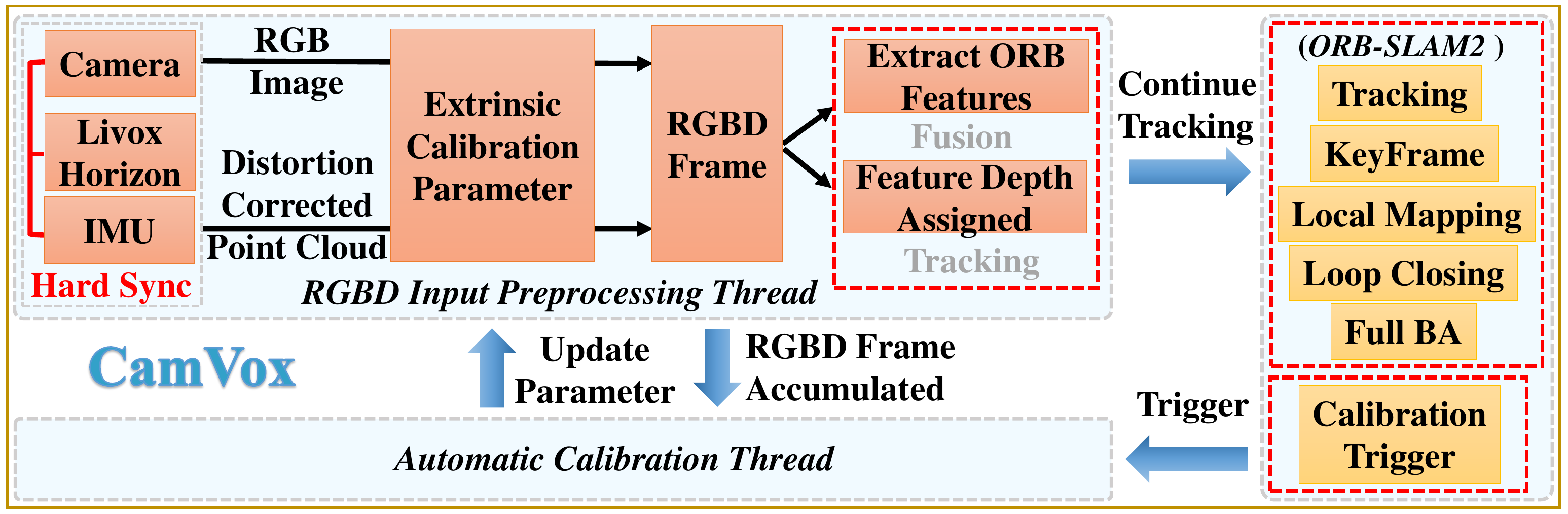}
    \caption{CamVox SLAM pipeline. In addition to the ORB-SLAM2 major threads, a RGBD input preprocessing thread is used to convert lidar and camera data to the RGBD format, and an automatic calibration thread can be automatically/manually triggered for camera and lidar extrinsic calibration, which is shown in Section C.}
    \label{fig:overview}
\end{figure}

The software pipeline runs on several parallel threads as shown in Fig. \ref{fig:overview}. In addition to the major threads from ORB-SLAM2, an additional RGBD input preprocessing thread is added to capture data from synchronized camera and lidar (IMU corrected) and process them into a unified RGBD frame. An automatic calibration thread can be triggered to calibrate the camera and lidar, which happens automatically when the robot is detected not moving or by human interaction. The calibrated result is then evaluated and output for potential parameter update.

\subsection{Preprocessing}

\begin{figure}[h]
    \centering
    \includegraphics[width=\linewidth]{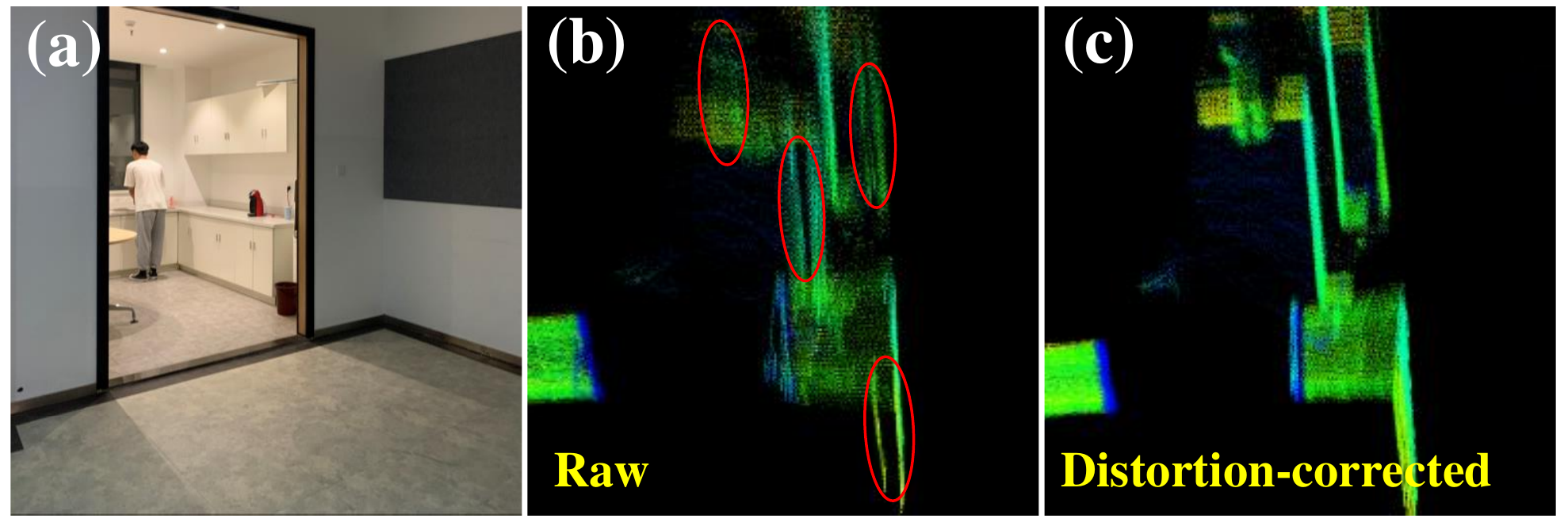}
    \caption{Example of motion distortion correction by IMU. (a) the RGB image of a corner scene; (b) the raw point cloud during a scan while the robot is moving; (c) the distortion corrected point cloud after processing with IMU data.}
    \label{fig:distortion_correction}
\end{figure}

The preprocessing thread takes the raw points from the lidar, corrected by the IMU and projected into a depth image according to the extrinsic calibration with camera. The RGB image is then combined with the depth image as the output of the RGBD frame, where the two images are formatted with equal size and pixel-wise corresponded as shown in Fig. \ref{fig:platform}c and Fig. \ref{fig:platform}d. Further tracking thread operations such as ORB feature extraction, keypoints generation are then performed based on the output. 
Since the lidar continuously scan the environment, each data point is obtained at a slightly different time stamp and needs correction from IMU. This is different from a camera, whereas an image is obtained at almost an instant. Fig. \ref{fig:distortion_correction} shows such an example when the point cloud is distorted while the robot is in a continuous motion. To correct this distortion, the robot motion is interpolated from the IMU pose at each lidar point output time, and transforms the lidar point to the lidar coordinate at the time when trigger signal starts, and that is also the time the camera image is captured. This transformation can be described by Eq. \eqref{eq:tsi} and Eq. \eqref{eq:psi}.

\begin{equation}\label{eq:tsi}
T_{si} = \frac{t_i-t_s}{t_e-t_s}T_{se}
\end{equation}

\begin{equation}\label{eq:psi}
P_i^s=T_{si}P_i^i 
\end{equation}

Where $t_s$ and $t_e$ are starting and ending timestamp of a lidar frame respectively. The timestamp of any 3D point in a lidar frame can be represented as $t_i$. According to the robot pose transformation matrix $T_{se}$ between $t_s$ and $t_e$ measured by IMU, $T_{si}$ between $t_s$ and $t_i$ can be calculated in \eqref{eq:tsi}. Finally, all lidar points are converted to the lidar coordinate system at $t_s$, which is shown in \eqref{eq:psi}.

\begin{figure}[h]
    \centering
    \includegraphics[width=\linewidth]{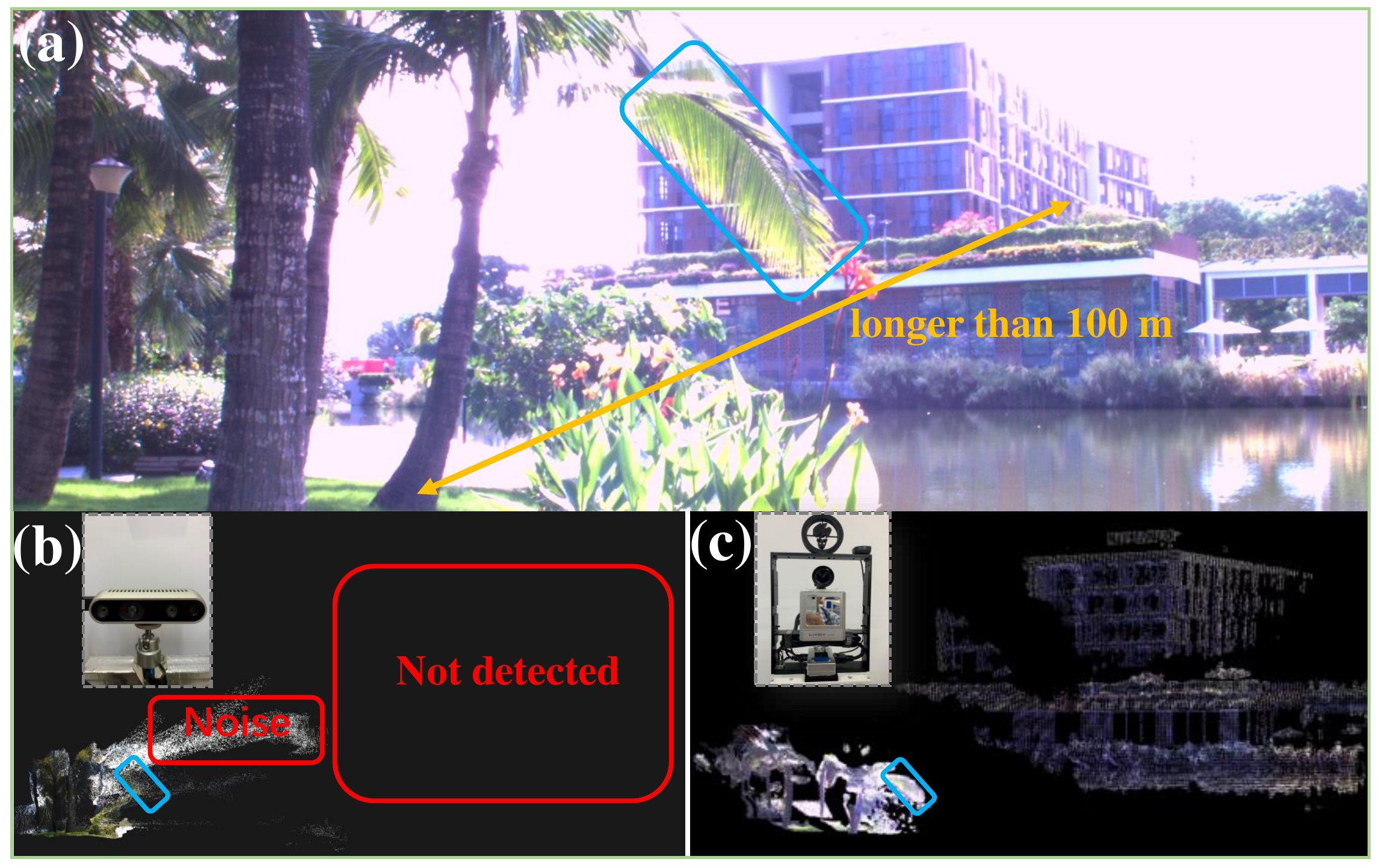}
    \caption{Student dormitory at SUSTech, captured by (a) RGB camera, (b) Realsense D435 and (c) CamVox RGBD output. The blue rectangles show the position of the coconut tree leaf at the three pictures. It is clear that typical RGBD camera are much worse than the CamVox RGBD output in detection distance and resistance to sunlight noise.}
    \label{fig:comparsion}
\end{figure}

Furthermore, with the help of long distance Livox lidar, we were able to detect reliably many depth-associated camera feature points beyond 100 meters. In comparison, the Realsense RGBD camera was not able to detect points beyond 10 meters and suffer from sunlight noises. This is clearly demonstrated in Fig. \ref{fig:comparsion}. As a result, in CamVox we can specify the close keypoints to those points with associated depth less than 130 meters. This is far more than the 40 times (ORB-SLAM2) the stereo baseline, which is on the order of 10 cm typically used in commercial stereo cameras.

\subsection{Calibration}
\begin{figure}[h]
    \centering
    \includegraphics[width=\linewidth]{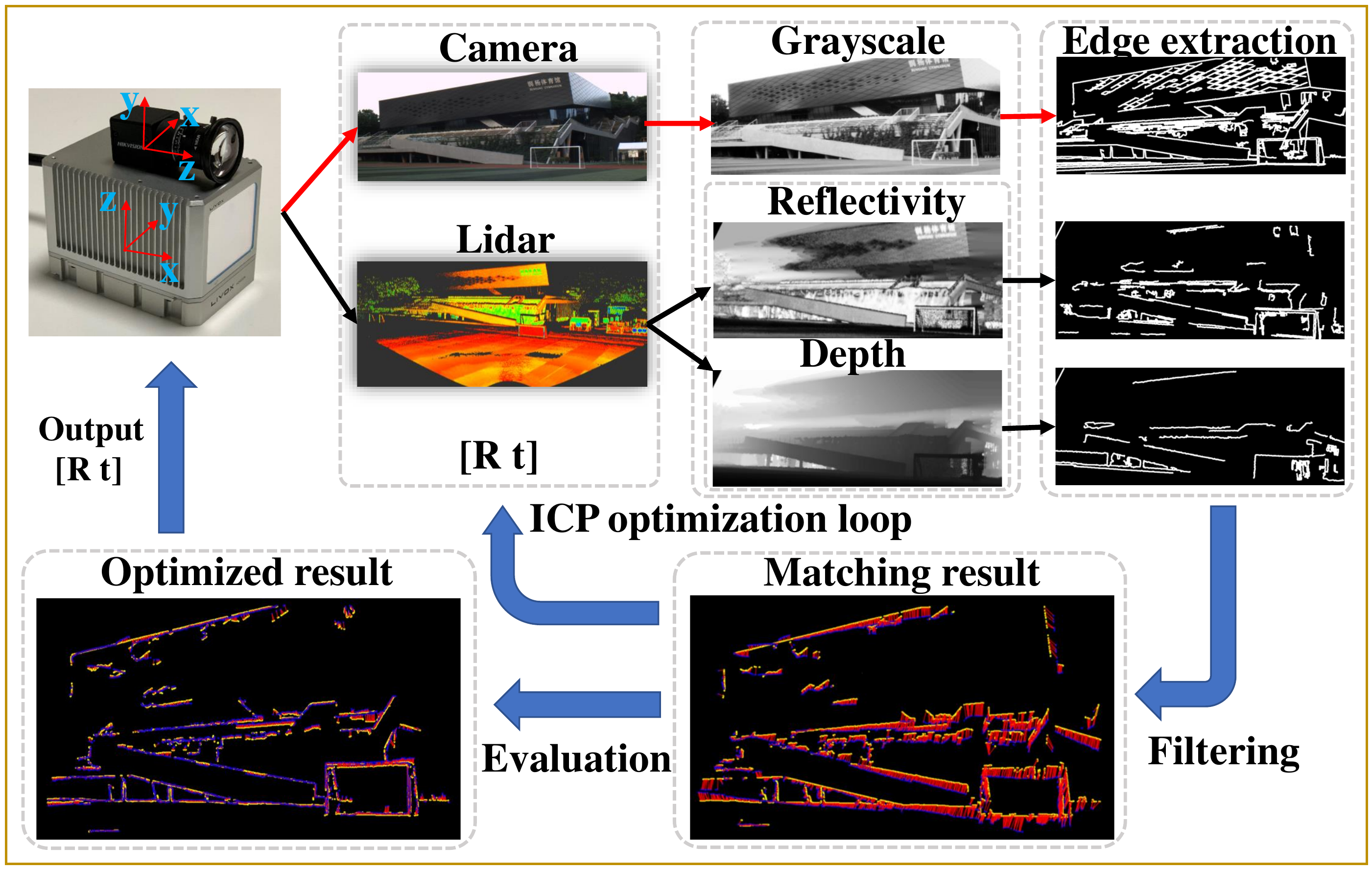}
    \caption{Calibration thread pipeline with examples. The captured data from camera and lidar (remaining still and accumulate for a few seconds) are processed with initial calibration parameters to form three images (grayscale, reflectivity and depth, the latter two are from lidar). Calibration is performed on the edges detected from these images until a satisfactory set of parameters is obtained.}
    \label{fig:correction_process}
\end{figure}

Calibration accuracy is vitally important in CamVox due to the long-range capability of the lidar. A small angular mismatch could result in a large absolute deviation at a large depth. Controlled calibration target such as checkerboards are not always available in the field and misalignment could happen after a random fixation failure or a collision. An automatic calibration method needs to be developed at an uncontrolled scene and update the parameters if a better calibration match is found. Thanks to the non-repeating nature of Livox lidars, as long as we could accumulate a few seconds of scanning points, the depth image could become as high resolution as a camera image (Fig. \ref{fig:correction_process}) and the correspondence to the camera image becomes easy to find. Therefore, we are able to do this calibration at almost all field scenes based on the scene information automatically. The triggering of this automatic calibration is set when the robot is detected to be stationary in order to eliminate the motion blur. We accumulate lidar points for a few seconds while remaining still. Camera image is also captured.

The overall calibration algorithm is structured as in Fig. \ref{fig:correction_process}. The dense point cloud is first projected onto the imaging plane by initial external parameters using both reflectivity and depth values, and contour extractions are then performed to compare with the camera image contour. The cost function is constructed by an improved ICP algorithm, which can be optimized by Ceres \cite{ceres-solver} to get the relatively more accurate external calibration parameters. Cost function from these new parameters is then evaluated against the previous values and a decision is made whether to update the extrinsic calibration parameter at input preprocessing thread.

Suppose the coordinate value of a point in the lidar coordinate system is $X=(x,y,z,1)^T$, the z coordiante value of a point in the camera coordinate system is $Z_{c}$ and the pixel position of the point in 2D image is $Y=(u,v,1)^T$. Given an initial external transform matrix from lidar to camera $T_{lidar}^{cam}$ and camera’s intrinsic parameter $(f_u,f_v,c_u,c_v)$, we can project the 3D point cloud to a 2D image by Eq. \eqref{eq:Prect} and Eq. \eqref{eq:Y}. 

\begin{equation}\label{eq:Prect}
P_{rect}=\left(\begin{array}{cccc}
f_{u} & 0 & c_{u} & 0 \\
0 & f_{v} & c_{v} & 0 \\
0 & 0 & 1 & 0
\end{array}\right)
\end{equation}

\begin{equation}\label{eq:Y}
Z_{c}Y = P_{rect}T_{lidar}^{cam}X
\end{equation}

After projecting lidar point cloud to 2D image, we do histogram equalization on all images and extract the edge using Canny edge detector \cite{canny1986computational}. The edges extracted from the depth image and the reflectivity image are combined because they are both from the same lidar but separate information. To extract more reliable edges, the following two kinds of edges are filtered out on those edge images. The first kind is the edge that is less than 200 pixels in length. The second kind of edge are the interior ones that are cluttered together. Finally, some characteristic edges that are present in both camera image and lidar image are obtained and edge matching are performed according to the nearest edge distance.

An initial matching result is shown at lower right of Fig. \ref{fig:correction_process}, where the orange line is the edge of the camera image, the blue line is the edge of the Horizon image, and the red line is the distance between the nearest points. Here we adopted the ICP algorithm \cite{besl1992method} and use K-D tree to speed up the search for the nearest neighbor point. However, sometimes in a wrong match very few points actually participated in the calculation of distance, the value of the cost function is trapped inside this local minimum. In this case, we improve the cost function in ICP by adding a degree of mismatching. The improved cost function is shown in Eq. \eqref{eq:CF}, where n is number of camera edge points that are within distance threshold with lidar edge points, m is the number of nearest neighbors, N is number of all camera points, b is weighing factor. We found a value of 10 for b is a good candidate to start as the default value. Note here that the cost function is an averaged value for each point and thus can be used to compare horizontally at different scenes.

\begin{equation}\label{eq:CF}
CF = \sum_{i=1}^{n} \sum_{i=1}^{m} \frac{\text{Distance}(P_{i}^{cam}, P_{ik}^{lidar})}{n\times m} + b \times \frac{N-n}{N}
\end{equation}

In optimizing this cost function, we adopted coordinate descent algorithms \cite{wright2015coordinate} and iteratively optimized (roll, pitch, yaw) coordinates by Ceres \cite{ceres-solver}. This seems to result in a better convergence. 

\section{Results}
In this section we present the evaluation results of CamVox. Specifically, we will first show the results of the automatic calibration. The effect of choosing depth threshold for close keypoints is also evaluated. Finally we evaluate the trajectory of CamVox as compared to some of the main stream SLAM frameworks and give a time analysis.

\subsection{Automatic calibration results}
\begin{figure}[h]
    \centering
    \includegraphics[width=\linewidth]{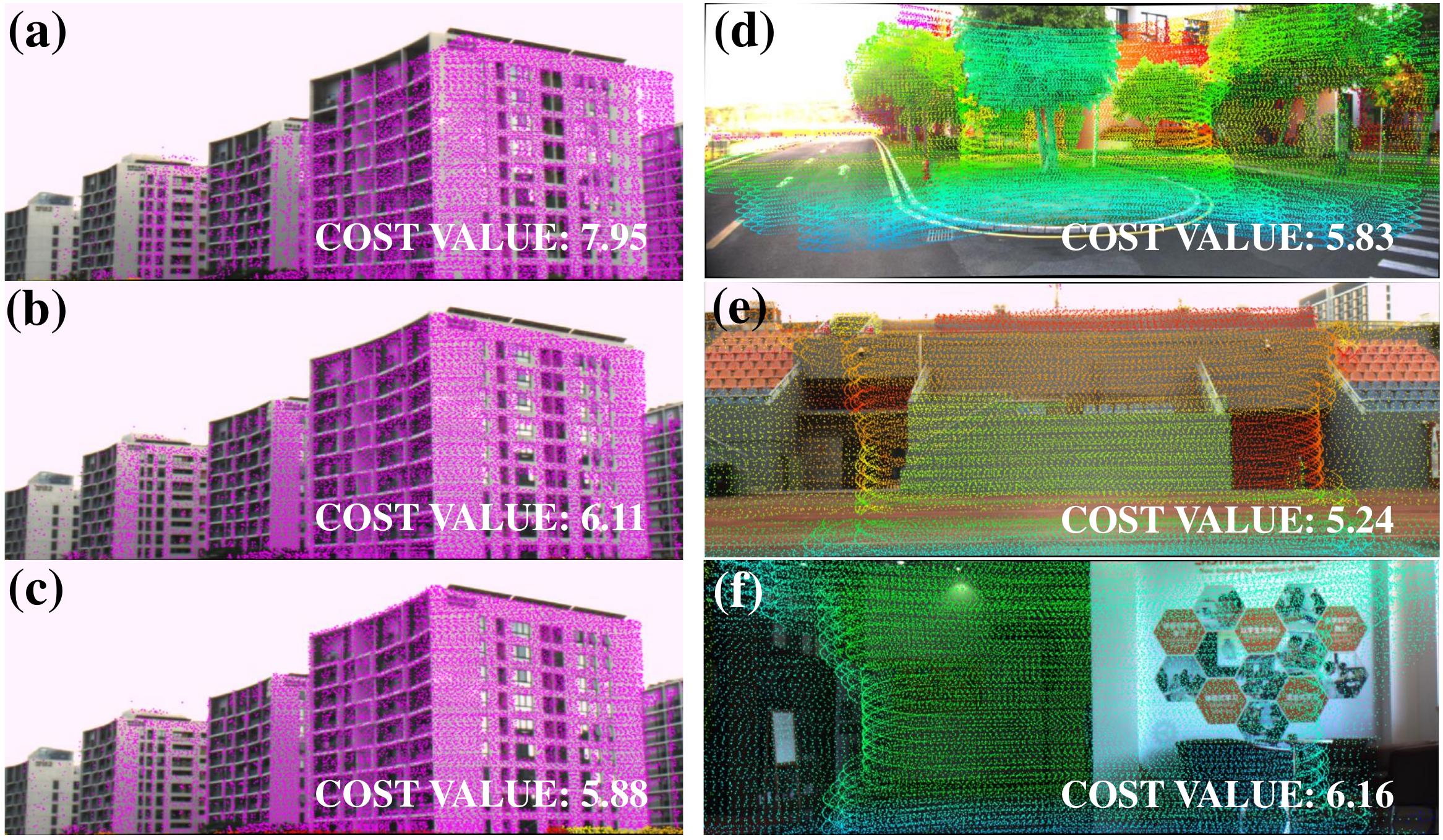}
    \caption{An example of RGB camera and point cloud overlay after calibration. (a) not calibrated. (b) automatically calibrated. (c) the best manual calibration. The automatic calibration algorithms is verified at various scenes, (d) outdoor road with natural trees and grasses, (e) outdoor artificial structures, (f) indoor underexposed structures.}
    \label{fig:correction_result}
\end{figure}

Our automatic calibration result is shown in Fig. \ref{fig:correction_result}. Shown in Fig. \ref{fig:correction_result}a is an overlay of lidar points onto the RGB image when the sensor and the camera are not calibrated (with a misalignment of more than 2 degrees). The cost function has a value of 7.95. The automatic calibration is triggered and calibrates the result as shown in Fig. \ref{fig:correction_result}(b), where the cost function value is 6.11, while the best possible calibration result done manually is shown in Fig. \ref{fig:correction_result}(c) with value 5.88. The automatic calibration delivers a very close result to the best manual calibration. Additionally, thanks to the image-like calibration scheme, the automatic calibration work robustly on most uncontrolled scenes. Several different scenes are evaluated in Fig. \ref{fig:correction_result}(d-f) and cost values are obtained around or below 6, which we regard as a relatively good value.

\subsection{Evaluation of depth threshold for keypoints}
\begin{figure}[h]
    \centering
    \includegraphics[width=\linewidth]{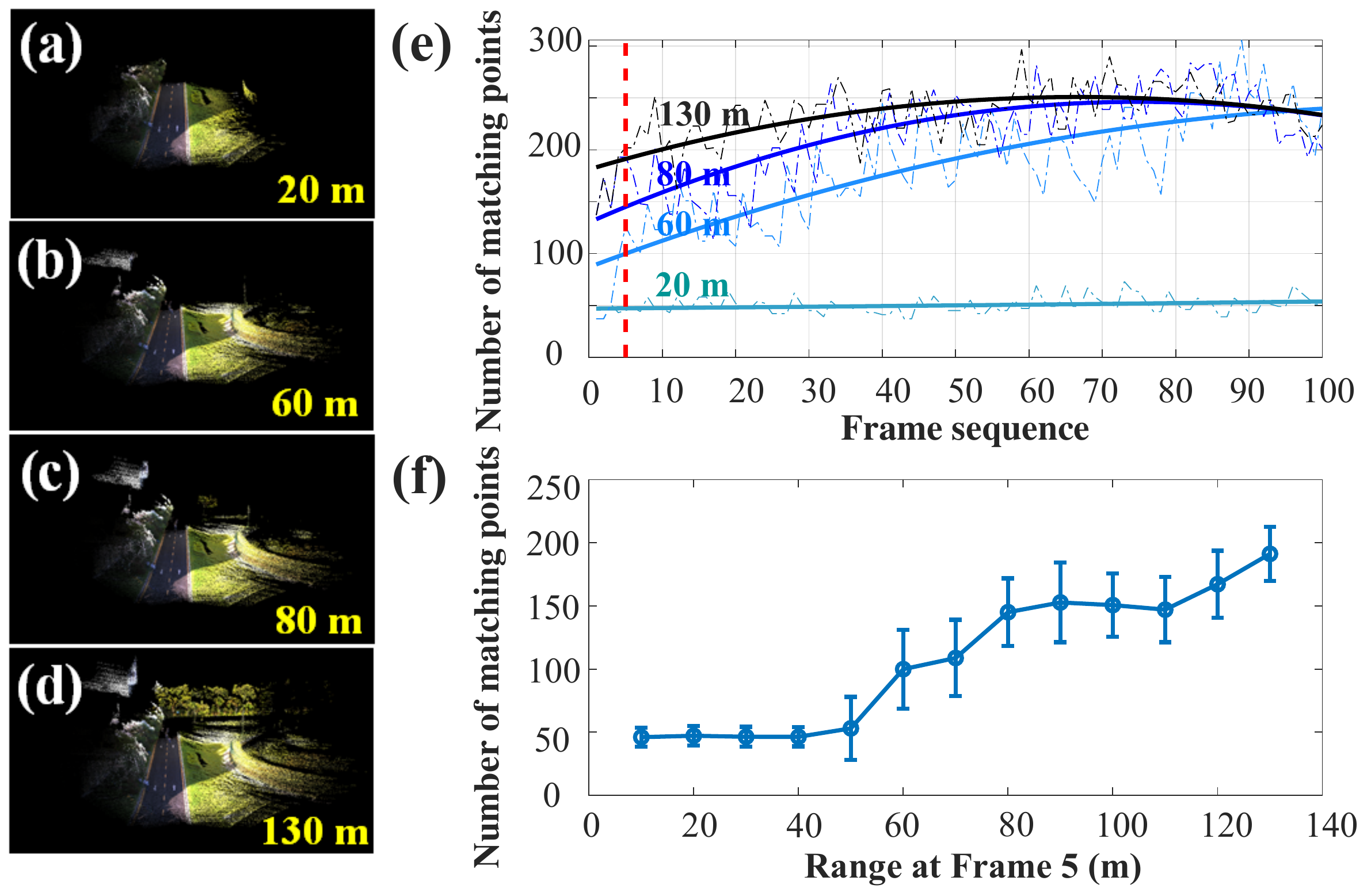}
    \caption{Evaluation of close keypoints threshold. (a-d) The reconstructed point cloud map by selecting different values of close keypoints threshold. (e) the number of matching points as a function of frame sequence (10 FPS) from start. (f) the number of matching points as a function of close keypoints threshold, evaluated at the beginning (5th frame) of SLAM process.}
    \label{fig:evalution}
\end{figure}

Because the lidar could detect 260 meters, there are many keypoints in the fused frame that we could characterize as close. These points help significantly in tracking and mapping. From Fig. \ref{fig:evalution}(a-d), by setting the close keypoints depth threshold from 20m to 130m, we see a significant increase in both mapping scale and number of mapped features. In Fig. \ref{fig:evalution}e We evaluated the number of matching points tracked as a function of time in the first 100 frames (10 FPS) after starting CamVox. An increase of feature numbers is observed as more frames is captured (Fig. \ref{fig:evalution}f 0.5 s after start), and the larger threshold obviously tracked more features initially that is helpful in starting a robust localization and mapping. With CamVox, we recommend and set the default value to 130 meters.

\subsection{Comparison of trajectory}
\begin{figure}[h]    \centering
    \includegraphics[width=\linewidth]{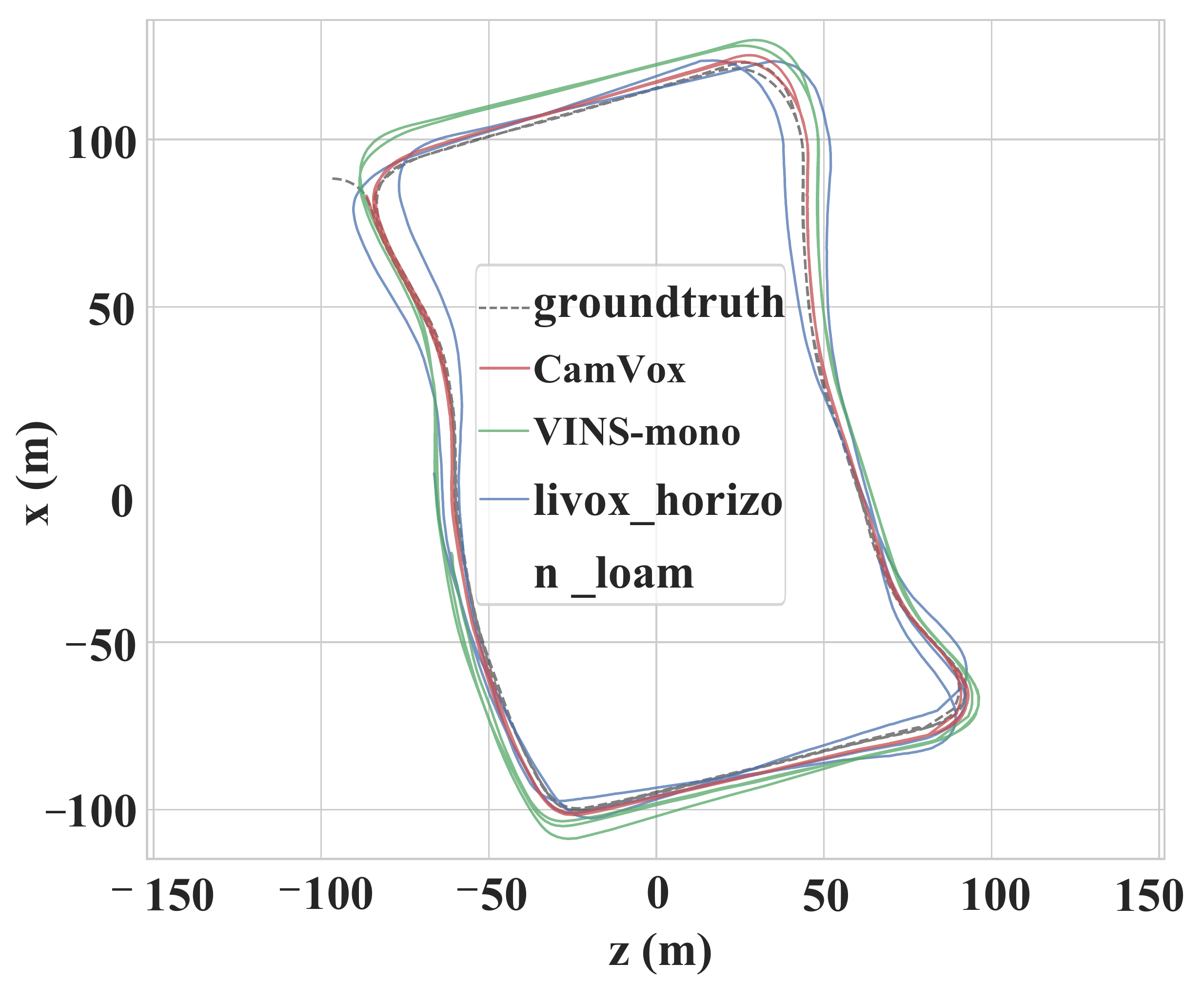}
    \caption{Trajectories from livox\_horizon\_loam, VINS-mono and  CamVox together with ground truth from SUSTech dataset.}
    \label{fig:trajectories}  
\end{figure}

The comparisons of the trajectories from CamVox, two mainstream SLAM framework and the ground truth are evaluated on our SUSTech dataset shown in Fig. \ref{fig:trajectories} and TABLE \ref{tab:error} using evo tools \cite{grupp2017evo} . Due to the accurate calibration, rich depth-associated visual features and their accurate tracking, CamVox system is very close to ground truth and significantly outperformed the other frameworks such as livox\_horizon\_loam and VINS-mono. 

\subsection{Timing results}
The timing analysis of the CamVox framework was performed as illustrated in TABLE \ref{tab:time}. The evaluation was performed on a onboard computer system Manifold 2C which has a 4-core Intel Core i7-8550U processor. With such a system, CamVox is able to perform in real-time. The automatic calibration takes about 58s to finish. Because calibration thread only runs occasionally while the robot is in a stationary state, and update of parameter could happen at a later time, such calculation time will not be an issue for real time performance.

\begin{table}[h]
\centering
\caption{Absolute Pose Error (APE) (unit: m)}
\label{tab:error}
\resizebox{\linewidth}{!}{%
\begin{tabular}{@{}cccc@{}}
\toprule
APE    & CamVox           & VINS-mono & livox\_horizon\_loam \\ \midrule
max    & \textbf{3.3}     & 27.2      & 9.9                 \\
mean   & \textbf{1.7}     & 6.7       & 6.2                 \\
median & \textbf{1.6}     & 6.1       & 6.5                 \\
min    & \textbf{0.2}     & 2.8       & 1.8                 \\
rmse   & \textbf{1.8}     & 7.5       & 6.5                 \\
sse    & \textbf{16066.5} & 50788.6   & 101223.7            \\
std    & \textbf{0.7}     & 3.5       & 1.9                 \\ \bottomrule
\end{tabular}%
}
\end{table}

\begin{table}[h]
\centering
\caption{Time Analysis}
\label{tab:time}
\resizebox{\linewidth}{!}{%
\begin{tabular}{@{}llll@{}}
\toprule
              & Framework      & CamVox          & ORB-SLAM2      \\ \midrule
Setting       & Dataset        & SUSTech         & TUM            \\
              & Resolution     & 1520$\times$568 & 640$\times$480 \\
              & Camera FPS     & 10 Hz            & 30 Hz           \\
              & ORB Features   & 1500            & 1000           \\ \midrule
Thread        & Calibration    & 58.16 s             & /              \\ 
              & Tracking       & 42.27 ms         & 25.58 ms        \\
              & Mapping        & 252.41 ms        & 267.33 ms       \\
              & Loop Closing   & 7821.22 ms          & 598.70 ms       \\ \midrule
RGBD          & IMU Corretcion & 0.89 ms             & /              \\
Preprocessing & Pcd2Depth      & 16.35 ms            & /              \\ \bottomrule
\end{tabular}%
}
\end{table}

\section{Conclusion and perspective}
To summarize, we have proposed CamVox as a new low-cost lidar-assisted visual SLAM framework, aiming to combine the best from both worlds, i.e., the best angular resolution from camera and the best depth and range from lidar. Thanks to the unique working principle of Livox lidar, an automatic calibration algorithm that could perform in uncontrolled scenes is developed. Evaluations of this new framework was carried out in automatic calibration accuracy, depth threshold for close keypoints classification and trajectory comparison. It could also run with real time performance on an onboard computer. We hope that this new framework could be useful for robotic and sensor research and an out-of-the-box low-cost solution for the community.

\addtolength{\textheight}{-12cm}   




\section*{ACKNOWLEDGMENT}

The authors thank colleagues at Livox Technology for helpful discussion and support.


\end{document}